\pdfoutput=1
\documentclass[11pt]{article}
\usepackage[preprint]{acl}
\usepackage{times}
\usepackage{latexsym}
\usepackage[T1]{fontenc}
\usepackage[utf8]{inputenc}
\usepackage{microtype}
\usepackage{inconsolata}
\usepackage{graphicx}
\usepackage{booktabs}
\usepackage{amsmath}
\graphicspath{{figs/}}

\newcommand{\cat}[1]{\textsc{#1}}

\title{Right Number, Wrong State? Measuring Cross-Jurisdiction\\Substitution in LLM Recall of State Policy}

\author{Jiayu Feng \\
  Harvard T.H. Chan School of Public Health \\
  \texttt{jiayufeng@hsph.harvard.edu}}

\begin{document}
\maketitle

\begin{abstract}
When an LLM answers a state-specific policy question wrongly, it may be hallucinating, or it may be returning a real value that holds in another state. We test this with a minimal-set design: the question wording is fixed and only the jurisdiction varies, across the 50 U.S.\ states and the District of Columbia (51 jurisdictions) and three exactly defined Medicaid income-eligibility quantities. Gold values come from an official data book and agree with an independent source in 101 of 102 checked cells. Under a pre-registered protocol, Claude Sonnet 5.5 and GPT-5.6 Sol reproducibly give another state's current value, identical across two independent repeats, for 10 and 25 of 153 items. The count depends on the reference set: crediting any wrong answer that equals another state's value yields 3 to 5 times as many reproducible substitutions as checking every number in the asked state's own records, because many of those answers also equal one of the asked state's own values under another convention or from an earlier year. Claims about cross-jurisdiction error need a complete same-state reference set. We will release the protocol, gold table, and all model outputs.
\end{abstract}

\section{Introduction}

Policy facts are \emph{jurisdiction-conditioned}: the same question (``What is the Medicaid income limit for pregnant women?'') has a different correct answer in each U.S.\ state. An LLM can return a real policy value bound to the wrong jurisdiction, a failure that a generic hallucination label hides. The failure is attractive to diagnose because it can be scored mechanically whenever the wrong answer equals another jurisdiction's gold value.

\begin{figure*}[t]
  \centering
  \includegraphics[width=\textwidth]{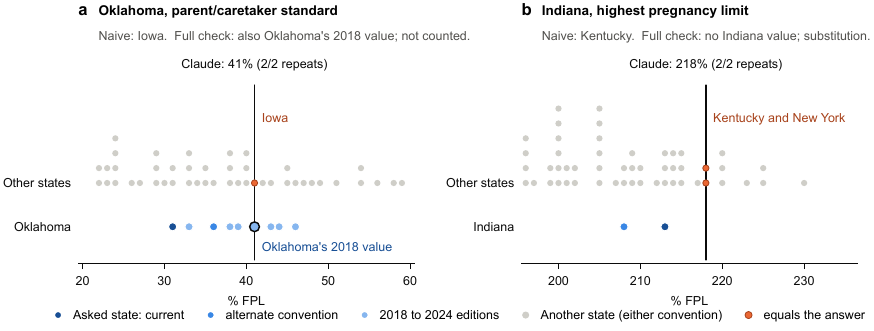}
  \caption{Two reproducible Claude answers on a number line; grey dots stack where several states share a value. (a) Oklahoma's 41\% equals Iowa's parent standard and Oklahoma's 2018 standard, so the primary protocol does not count it. (b) Indiana's 218\% equals Kentucky's limit (and New York's without the disregard) and no Indiana value from 2018 to 2025, so it remains a substitution.}
  \label{fig:design}
\end{figure*}

Numeric policy facts admit several contemporaneous representations (e.g., with or without a 5-point income disregard) and change over time, so a wrong answer that ``matches another state'' often also equals the asked state's own value under another reading (Figure~\ref{fig:design}). Our evaluation asks for one exactly defined policy quantity per question, in each of the 51 jurisdictions, and scores answers against an authoritative gold table. A pre-registered taxonomy checks the asked state's own values (current, under the other disregard convention, or from an earlier year) before federal anchors, other states' values, and unsupported numbers, and counts an item only when independent repeats agree. Under this protocol, cross-jurisdiction substitution appears in both model families we test, and its count depends on the reference set.

\section{Related work}

\textbf{Factuality evaluation.} Factuality benchmarks ask whether an answer is supported or correct: FActScore checks atomic claims against a knowledge source \citep{min-etal-2023-factscore}, and SimpleQA grades short answers against a single reference \citep{wei2024simpleqa}. In both, a wrong answer counts only as an error. We also ask whether a wrong answer's value belongs to a competing entity; that check requires knowing every value the asked entity could legitimately have.

\textbf{Entity- and context-conditioned recall.} Language models store relational facts that cloze queries can retrieve \citep{petroni-etal-2019-language}, recall is weaker for less popular entities \citep{mallen-etal-2023-trust}, and geospatial predictions track ground truth while carrying systematic geographic biases \citep{manvi2024geographic}. Swapping the entity in a passage reveals reliance on memorized answers \citep{longpre-etal-2021-entity}. SituatedQA \citep{zhang-choi-2021-situatedqa} and temporal probing \citep{dhingra-etal-2022-time} show that answers depend on time and place, and stale gold labels distort factuality scores \citep{jiang2026benchmarks}. Holding the query fixed while an explicit jurisdiction varies over a complete set of entities that share one attribute lets a wrong value be matched to every entity that holds it, and stale \emph{model} answers then form their own error class.

\textbf{Numbers and value collisions.} NLP systems rarely give numbers special treatment \citep{thawani-etal-2021-representing}, and language models bind attributes to entities in context through learned internal mechanisms \citep{feng2024bind}. Knowledge overshadowing proposes that dominant knowledge can obscure less prominent knowledge \citep{zhang2025law}, a hypothesis our leakage test probes. Numeric attributes make attribution hard: with 30 to 36 distinct values shared by 51 jurisdictions, a wrong number often equals some other entity's value by coincidence, so a match alone is weak evidence of substitution.

\textbf{Jurisdiction in legal and benefits QA.} Legal hallucination rates vary by jurisdiction and place \citep{dahl2024large,curran2025place}; Housing Statute QA asks the same yes/no questions across 50+ jurisdictions \citep{zheng2025reasoning}; CrossLex holds facts fixed across three legal systems \citep{yang2026crosslex}; Public Benefits Bench grades state-specific Supplemental Nutrition Assistance Program (SNAP) guidance with rubrics \citep{vals2026benefits}; and \citet{taranukhin2026warrant} count jurisdictional applicability as part of legal warrant. None attributes wrong answers to a specific other jurisdiction, because binary or rubric-graded answers carry no source identity.

\begin{figure*}[t]
  \centering
  \includegraphics[width=\textwidth]{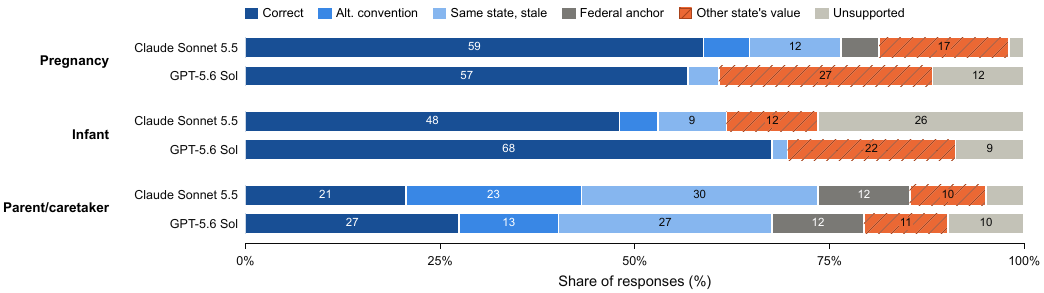}
  \caption{Outcomes per model and quantity, as percentages of 102 responses (51 jurisdictions $\times$ 2 repeats); segments below 8\% are unlabeled. Blue shades are the asked state's own values (current, alternate convention, or an earlier edition); hatched orange is another state's current value.}
  \label{fig:outcomes}
\end{figure*}

\section{Task and gold data}

\textbf{Quantities.} We ask, ``As of July 2025, what is [quantity] in [state]?'' for three Medicaid income-eligibility quantities, each defined in the question itself: \emph{pregnancy}, the highest limit for pregnancy coverage under Medicaid, the Children's Health Insurance Program (CHIP), or the CHIP unborn-child option, including the 5-point disregard; \emph{infant}, the highest Medicaid/CHIP limit for infants under one, including the disregard; and \emph{parent}, the published standard of the parent/caretaker eligibility group, excluding the disregard (the disregard applies only to an applicant's highest eligibility group, so the standard itself is the policy quantity, although KFF reports parent limits with the disregard added; \citealp{kff2026parents}). Answers are percentages of the federal poverty level, returned as JSON.

\textbf{Gold.} Values are derived deterministically from MACStats Exhibits 35 and 36 of the Medicaid and CHIP Payment and Access Commission (MACPAC) \citep{macpac2026macstats} for the 51 jurisdictions (153 items; 30 to 36 distinct values per quantity, modal share $\le$12\%). The derived pregnancy and infant values agree with KFF's January 2026 tables \citep{kff2026pregnancy,kff2026children} in 51/51 and 50/51 states; the one difference follows a MACPAC footnote. Each item has exactly one \emph{alternate convention}: the same figure with the disregard handled the other way ($-$5 for pregnancy and infant, $+$5 for parent). It is a reporting convention for the same eligibility limit. Each item also has a set of \emph{stale} values computed identically from every MACStats edition 2018 to 2024, so that the asked state's own earlier values are not misclassified as cross-jurisdiction substitutions.

\section{Protocol}

\textbf{Models.} We evaluate Claude Sonnet 5.5 and GPT-5.6 Sol. Sonnet runs at the lowest effort and its default temperature, with thinking restricted to tool calls and no tools provided, so each output is the JSON answer alone. GPT uses reasoning effort \emph{none} and temperature 0. Both run closed-book with no tools or retrieval, strict JSON output, one fresh request per item, and two independent repeats.

\textbf{Taxonomy.} Each response receives the first matching category: \cat{Unparsable}; \cat{Correct}; \cat{Alternate} convention; same-state \cat{Stale}; \cat{Federal anchor} (100/133/138\%); \cat{Other-state} (equals another state's current value, either convention); \cat{Unsupported}. Same-state explanations take precedence, so cross-state credit is conservative. An item is \emph{reproducible} when both repeats give the identical value; reproducible errors are reproducible items other than \cat{Correct} and \cat{Alternate}. Our primary measure is \emph{reproducible cross-jurisdiction substitution}: reproducible items classified \cat{Other-state}. Response-level rates are descriptive, and the attribution levels in Figure~\ref{fig:protocol} are sensitivity analyses.

\textbf{Pre-registration.} Before collecting the reported data, we fixed the prompt, the gold table (including the 2018 to 2024 stale values), the taxonomy and scorer, the number of repeats, the primary analyses, and a criterion for continuing the study: a non-Claude model has $\ge$8 reproducible \cat{Other-state} items making up $\ge$10\% of its reproducible errors, or shows significant \emph{known-fact leakage} (a value the model gives correctly for its own state appears as the answer for another state; Mantel-Haenszel over the $n=35$ source states with a unique value). Claude was used during design, so the criterion applies to non-Claude models and the reported Claude data come from a new run collected after the protocol was locked. The protocol comparison in Figure~\ref{fig:protocol} and the stability-free comparison were specified after the GPT run and before the new Claude run. A third model added to the protocol by amendment was stopped after 9 of 306 calls because the service was unavailable; none of its responses was scored.

\begin{table}[t]
\centering\small
\setlength{\tabcolsep}{3pt}
\resizebox{\columnwidth}{!}{%
\begin{tabular}{llcrrrrrr}
\toprule
 & & Acc.\ [95\% CI] & Alt & Stale & Anch & \textbf{Oth} & Uns & Unst \\
\midrule
Sonnet & Preg. & .59 [.46,.72] & 1 & 6 & 2 & 4 & 0 & 11 \\
 & Infant & .48 [.36,.60] & 1 & 4 & 0 & 3 & 9 & 15 \\
 & Parent & .21 [.10,.32] & 9 & 12 & 6 & 3 & 2 & 9 \\
\midrule
GPT & Preg. & .57 [.43,.71] & 0 & 2 & 0 & 12 & 6 & 3 \\
 & Infant & .68 [.55,.79] & 0 & 1 & 0 & 9 & 4 & 4 \\
 & Parent & .28 [.16,.39] & 6 & 13 & 6 & 4 & 5 & 4 \\
\bottomrule
\end{tabular}}
\caption{Accuracy over 102 responses per cell, with 95\% confidence intervals (CI) from a jurisdiction-level bootstrap; other columns count the 51 items by reproducible outcome (Unst = repeats differ). No response was unparsable.}
\label{tab:main}
\end{table}

\section{Results}

Both models answer the pregnancy and infant questions correctly in 48 to 68\% of responses, but the parent question in only 21 to 28\% (Table~\ref{tab:main}).

Error types differ by quantity and by model (Figure~\ref{fig:outcomes}). Stale values dominate parent errors (12 and 13 reproducible items): dollar-denominated standards drift in percentage terms every year, and models return earlier years' figures. Federal anchors appear mostly in parent answers. At the response level, other-state answers make up 12.7\% of Claude's responses and 19.9\% of GPT's, concentrated in GPT's pregnancy and infant answers. Identifiable GPT cases include New Mexico infants answered with New York's 405\% and Louisiana pregnancy answered with Maryland's 259\% (Maryland's value without the disregard).

\textbf{Substitution is reproducible in both models.} GPT has 25 reproducible \cat{Other-state} items (40\% of its 62 reproducible errors) and Claude has 10 (20\% of 51); GPT meets the continuation criterion. Known-fact leakage is not significant for either model (pooled $p=.41$ and $.66$). GPT's repeats agree on 93\% of items and Claude's on 77\%. Counting an item when \emph{either} repeat gives another state's value yields 34 items for GPT and 27 for Claude (exact McNemar test between the models, $p=.38$).

\begin{figure}[t]
  \centering
  \includegraphics[width=\columnwidth]{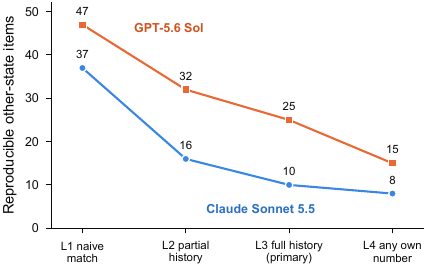}
  \caption{Reproducible other-state items under four attribution protocols. L1 credits any wrong answer equal to another state's value; L2 adds the alternate convention and three data-book editions; L3 (primary) uses every edition 2018 to 2024; L4 also discounts any number in the asked state's MACStats rows.}
  \label{fig:protocol}
\end{figure}

\textbf{The substitution count depends on the reference set} (Figure~\ref{fig:protocol}). Crediting any wrong answer that equals another state's value yields 37 reproducible substitutions for Claude and 47 for GPT. The primary protocol yields 10 and 25, and discounting every number in the asked state's data-book rows yields 8 and 15. The difference is answers that also equal the asked state's own values. Claude's Oklahoma parent answer of 41\% equals Iowa's current value and Oklahoma's 2018 standard (Figure~\ref{fig:design}); its Indiana parent/caretaker answer of 16\% equals Texas's value with the disregard and Indiana's 2022 standard. GT-FENG \citep{feng2026gtfeng} labels such answers source-ambiguous. Extending a partial value history (L2) to the full 2018 to 2024 history (L3) removes 6 of Claude's 16 and 7 of GPT's 32 reproducible substitutions.

\section{Conclusion}

Both model families reproducibly return other jurisdictions' values: under the primary protocol, GPT-5.6 Sol does so for 25 of 153 items and Claude Sonnet 5.5 for 10, and under the most conservative check for 15 and 8. How many wrong-state answers an evaluation finds depends on its reference set. Without the asked jurisdiction's own records (its alternate conventions, its value history and the other figures in its data-book rows), the count here is 3 to 5 times as high. Benchmarks of facts that vary by jurisdiction, date, or reporting convention face the same problem. Writing the exact quantity into each question removes doubt about which convention is scored, and listing every value the asked entity has held shows which wrong answers also equal the asked entity's own values before another entity is credited. Repeats screen out one-off guesses, and reporting counts at more than one attribution level shows readers how much of a wrong-entity rate comes from the protocol itself.

\section*{Limitations}

The study covers one policy domain, three quantities, two models, and one prompt wording. It says nothing about other domains or prompts, or about retrieval-augmented systems, where tool use may dominate \citep{vals2026benefits}. MACPAC compiles state data, and its parent standards depend on its own dollar-to-percentage conversions, which no second source checks. Values before 2018 were not checked because earlier tables use a different layout, so a few remaining \cat{Other-state} items may still be older same-state values.

With only 35 source states holding a unique value, the leakage test has low power, and its null result does not show that the mechanism is absent. Repeat stability differs between the models (temperature 0 versus an unadjustable default), which affects counts that require agreement across repeats; we report response-level rates and the stability-free comparison for this reason. Claude's counts also vary between runs, so small differences should not be over-read. A match with another state's value is consistent with retrieval of that state's fact but does not prove it, and widely shared values such as 205\% cannot be traced to any one state.

\section*{Ethical considerations}

All data are public aggregate policy statistics; no personal data or human subjects were involved. Wrong eligibility thresholds could mislead applicants; we measure model behavior and do not assess downstream harm.

\bibliography{refs}

\end{document}